%% file: neurips_2025.tex
\documentclass{article}

\PassOptionsToPackage{authoryear,square,semicolon}{natbib}

\usepackage[preprint]{neurips_2025}

\usepackage[utf8]{inputenc} % allow utf-8 input
\usepackage[T1]{fontenc}    % use 8-bit T1 fonts
\usepackage{hyperref}       % hyperlinks
\usepackage{url}            % simple URL typesetting
\usepackage{booktabs}       % professional-quality tables
\usepackage{amsfonts}       % blackboard math symbols
\usepackage{nicefrac}       % compact symbols for 1/2, etc.
\usepackage{microtype}      % microtypography
\usepackage{xcolor}         % colors
\usepackage{amsmath}
\usepackage{wasysym}   % 提供半对勾符号
\usepackage{colortbl}
\usepackage{fontawesome} 
\usepackage{enumitem}
\usepackage{wrapfig}

\usepackage{enumitem}
\usepackage{graphicx}
\usepackage{multirow}
\usepackage{subfigure}
\usepackage{subcaption}
\usepackage{array}
\usepackage{algorithmic}
\usepackage[ruled, linesnumbered]{algorithm2e}
\usepackage{xcolor}
\definecolor{codegreen}{rgb}{0,0.6,0}

\usepackage{amsthm}
\usepackage{booktabs}
\usepackage{makecell}
\usepackage{graphicx}
\usepackage{pifont}

\usepackage{hyperref}
\usepackage{xurl}      % 让长 URL 可断行
\hypersetup{breaklinks=true}
\newcommand\cmark {\textcolor{green}{\ding{52}}}
\newcommand\xmark {\textcolor{red}{\ding{55}}}
\definecolor{lightgray}{gray}{0.95}

\newcommand{\ourmethod}{BrowseComp-V$^3$}

\newcommand{\AuthorSep}{\hspace{0.55em}}   % 作者之间间距
\newcommand{\AffilSep}{\hspace{1.0em}}     % 单位之间间距
\newcommand{\BlockGapA}{0.35em}            % 作者 -> 单位 的竖直间距
\newcommand{\BlockGapB}{0.25em}            % 单位 -> 注释 的竖直间距

\title{BrowseComp-V$^3$: A Visual, Vertical, and Verifiable Benchmark for Multimodal Browsing Agents}

\author{%
\parbox{\textwidth}{\centering\footnotesize
Huanyao Zhang$^{1,\diamondsuit,*}$,\AuthorSep
Jiepeng Zhou$^{2,\diamondsuit}$,\AuthorSep
Bo Li$^{1,\diamondsuit}$,\AuthorSep
Bowen Zhou$^{1,\diamondsuit}$,\AuthorSep
Yanzhe Shan$^{3,\diamondsuit}$,\AuthorSep
Haishan Lu$^{1}$,\AuthorSep
Zhiyong Cao$^{4}$,\AuthorSep
Jiaoyang Chen$^{5}$,\AuthorSep
Yuqian Han$^{1}$,\AuthorSep
Zinan Sheng$^{1}$,\AuthorSep
Zhengwei Tao$^{1}$,\AuthorSep
Hao Liang$^{1}$,\AuthorSep
Jialong Wu$^{1}$,\AuthorSep
Yang Shi$^{1}$,\AuthorSep
Yuanpeng He$^{1}$,\AuthorSep
Jiaye Lin$^{6}$,\AuthorSep
Qintong Zhang$^{1}$,\AuthorSep
Guochen Yan$^{1}$,\AuthorSep
Runhao Zhao$^{1}$,\AuthorSep
Zhengpin Li$^{1}$,\AuthorSep
Xiaohan Yu$^{7}$,\AuthorSep
Lang Mei$^{7}$,\AuthorSep
Chong Chen$^{7,\dagger}$,\AuthorSep
Wentao Zhang$^{1,\dagger}$,\AuthorSep
Bin Cui$^{1,\dagger}$\\[\BlockGapA]
$^1$PKU\AffilSep
$^2$HKUST(GZ)\AffilSep
$^3$OUC\AffilSep
$^4$CASIA\AffilSep
$^5$HITSZ\AffilSep
$^6$THU\AffilSep
$^7$Huawei Cloud BU\\[\BlockGapB]
{\scriptsize $^\diamondsuit$ Core Contributor \quad $^*$ Project Leader \quad $^\dagger$ Corresponding author}
}
}

\begin{document}

\maketitle

\begin{abstract}
Multimodal large language models (MLLMs), leveraging their increasingly advancing autonomous planning and tool use capabilities, are evolving into intelligent agents capable of performing web browsing for multimodal deep search. 
However, existing benchmarks remain limited in terms of task complexity, information searchability, and evaluation dimensions, thereby hindering comprehensive assessments of multimodal browsing agents' deep search capabilities in open-world environments.
To bridge these gaps, we present \ourmethod, a novel benchmark comprising 300 meticulously hand-crafted, challenging questions across diverse domains. 
By emphasizing deep, multi-level, and cross-modal multi-hop reasoning, we ensure that these tasks necessitate the use of web browsing tools and cannot be resolved solely through the model's parametric knowledge.
Moreover, we strictly enforce the public searchability of all supporting evidence and incorporate an expert-validated, subgoal-driven process evaluation mechanism, thereby enabling fine-grained characterization of search behaviors and systematic analysis of capability boundaries.
Beyond the dataset, we provide OmniSeeker, a general multimodal browsing agent framework, and conduct a comprehensive evaluation on MLLMs. The results demonstrate that even state-of-the-art models, such as GPT-5.2, achieve only 36\% accuracy.
Further analysis reveals critical bottlenecks in existing models regarding multimodal information integration and fine-grained perception, highlighting a fundamental lack of native multimodal reasoning capabilities.
% Project page: \url{https://halcyon-zhang.github.io/BrowseComp-V3}
\end{abstract}

%%%%%%%%%%%%%%%%%%%%%%%%%%%%%%%%%%%%%%%%%%%%%%%%%%%%%%%%%%%%

\section{Introduction}
\label{sec:intro}
% 1. bg related works

Multimodal large language models~\citep{openai2025gpt52, pichai2025gemini3, bai2025qwen3vltechnicalreport, llama4, li2024llava, shi2025mavors} have demonstrated substantial performance gains across complex tasks. By integrating linguistic comprehension, visual perception, and tool-use capabilities, these models are increasingly evolving into autonomous agents capable of independent exploration and decision-making. Consequently, an increasing body of research is exploring how MLLMs can leverage external search and browsing tools to address multimodal  deepsearch challenges in open-world environments~\citep{oaidr, geminidr, wu2025mmsearchr1, geng2025webwatcher, huang2026vision}.

Despite the rapid evolution of model capabilities, benchmarks for multimodal browsing and deep search remain noticeably underdeveloped. Existing studies~\citep{jiang2024mmsearch, geng2025webwatcher, li2025mmbrowsecomp, tao2025mmsearchplus} frequently exhibit shortcomings in task complexity, information searchability, and evaluation dimensions, hindering fair, holistic, and reproducible assessments of multimodal browsing agents. Existing methods still exhibit certain limitations: \textbf{i) Insufficient Task Complexity.} Early benchmarks~\citep{ cheng2025simplevqa, geng2025webwatcher} are predominantly confined to shallow retrieval within two hops, with visual information concentrated in the initial stage. Consequently, they fail to reflect the intricacies of real-world, deep multimodal search scenarios. \textbf{ii) Inaccessibility of Key Information.} The core evidence in subsequent benchmarks~\citep{fu2025livevqa,li2025mmbrowsecomp, tao2025mmsearchplus} is often derived from sources that are not publicly searchable by tools, such as videos or proprietary documents, which undermines the reproducibility and fairness. \textbf{iii) Narrow Evaluation Dimensions.} Existing studies~\citep{jiang2024mmsearch, li2024DynVQA} primarily focus on the accuracy of the final answer but lack a systematic characterization of the reasoning process. This makes it challenging to diagnose specific failure modes or define the capability boundaries of the models.

To address these gaps, we present \ourmethod, a novel benchmark specifically designed to evaluate multimodal deep browsing and search capabilities. 
\ourmethod~comprises 300 carefully curated, highly complex questions spanning 24 distinct sub-domains, which systematically assess multimodal browsing agents in open-world settings.
A key feature of our work is the emphasis on deep, multi-level, and cross-modal reasoning, where critical evidence is strategically interleaved across textual and visual modalities within and across web pages. This design effectively precludes "shortcut" successes derived solely from text-based heuristics or models' reliance on internal parametric knowledge.
Furthermore, we ensure that all critical evidence is accessible via standard public search engines and provide manually annotated gold-standard search trajectories to guarantee fairness and reproducibility.
Finally, we introduce expert-validated intermediate sub-goals for each task, enabling fine-grained evaluation of the search process to precisely identify the capability boundaries and failure modes of the evaluated models. Our primary contributions are summarized as follows:
\begin{itemize}[leftmargin=0.5cm] 
    \item We present \ourmethod, which, to the best of our knowledge, represents the first multimodal deep search benchmark to concurrently feature extensive search depth, public search accessibility, and process-oriented evaluation mechanisms.
    \item We systematically define and categorize multimodal deep search scenarios. Through process-oriented evaluation, we provide a more comprehensive characterization of multimodal browsing agents' capabilities and limitations.
    \item We develop OmniSeeker, a unified multimodal browsing agent framework. By integrating diverse web search and visual perception tools, OmniSeeker rivals the performance of state-of-the-art closed-source systems and substantially enhances open-source models' performance on multimodal deep search tasks.
\end{itemize}

\input{table/comparison}

\section{Related Work}

\subsection{Multimodal Large Language Models}
Multimodal large language models~\citep{openai2025gpt52, pichai2025gemini3, bytedance2025seed18, bai2025qwen3vltechnicalreport} have demonstrated remarkable proficiency across a diverse spectrum of tasks, such as VQA~\citep{fu2025video, cheng2025simplevqa, zhang2025debiasing, wu2025mmsearchr1}, grounding~\citep{kazemzadeh2014referitgame}, OCR~\citep{masry2022chartqa, shi2025realunify, mathew2021docvqa, shi2025mme}, and multimodal reasoning~\citep{lu2023mathvista, wang2025monet, wang2024measuring}. 
Nevertheless, MLLMs inherently struggle with the real-time acquisition of up-to-date information, posing substantial hurdles when addressing knowledge-intensive or information-retrieval queries. 
Consequently, contemporary research has pivoted toward tool-augmented frameworks to empower MLLMs as autonomous agents, capable of dynamically retrieving and incorporating external knowledge.

\subsection{Tool-Enhanced Browsing Agents}
Driven by the escalating tool-calling proficiency of LLMs/MLLMs~\citep{guo2025deepseekr1, yang2025qwen3, openai2025gpt52}, tool-enhanced browsing agents have emerged as a pivotal research frontier. 
To enable precise retrieval and reasoning in dynamic web environments, recent studies advocate leveraging supervised fine-tuning and reinforcement learning to enhance agents' reasoning and decision-making capabilities~\citep{jin2025search, li2025websailor, wu2025webdancer, tao2025webshaper, song2025r1searcher, zheng2025deepresearcher}.
This paradigm, initially validated in textual agents, has been rapidly extended to the multimodal domain~\citep{wu2025mmsearchr1, mei2025aiplan, hong2025deepeyesv2, huang2026vision}. This evolution has significantly expanded the search depth and adaptive boundaries of agents when navigating complex tasks.

\subsection{Multimodal Browsing Benchmarks}
Traditional multimodal browsing benchmarks typically decouple visual understanding from text retrieval and focus on simple two-hop retrieval tasks~\citep{chen2023infoseek, mensink2023Enc-VQA, cheng2025simplevqa, geng2025webwatcher}.
As visual agents advance, performance on such tasks has largely saturated.
BrowseComp~\citep{wei2025browsecomp} evaluates text-only agents in open-world settings by requiring large-scale web navigation, offering valuable guidance for multimodal task construction. Inspired by this paradigm, recent benchmarks such as MM-BrowseComp~\citep{li2025mmbrowsecomp} and MMSearch-Plus~\citep{tao2025mmsearchplus} incorporate multi-hop designs and fine-grained visual reasoning to enhance reasoning depth.
However, existing benchmarks still suffer from key limitations: critical information often resides in videos or non-searchable documents, tool support is insufficient, and evaluation primarily measures final-answer correctness while overlooking the quality of reasoning.
To bridge this gap, we propose \ourmethod, ensuring all critical information comes from publicly accessible resources during task design. We also introduce the Process Score metric to evaluate multimodal browsing agents comprehensively.

\begin{figure*}
    \centering
    \includegraphics[width=\textwidth]{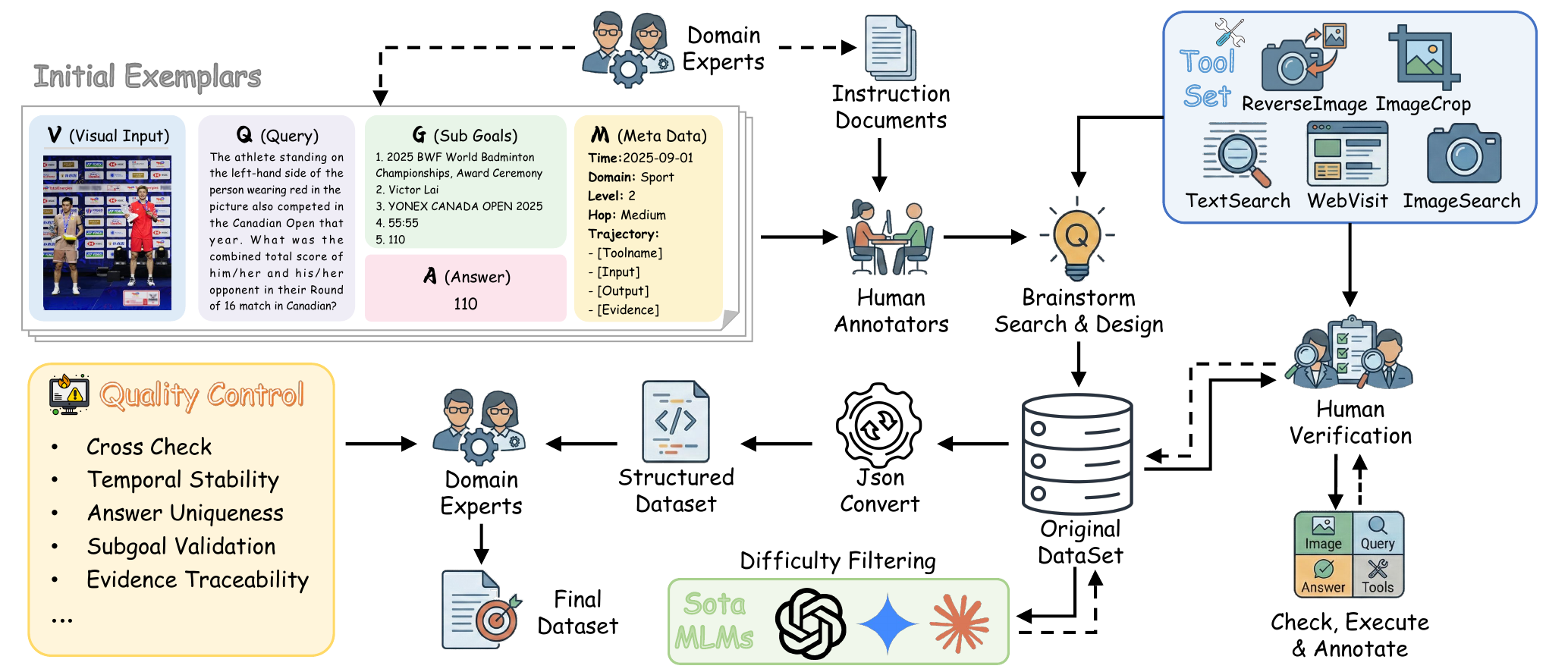}
    \caption{An overview of the data construction process of \ourmethod.}
    \label{fig:overview}
\end{figure*}

\section{The BrowseComp-V$^3$ Dataset}
\ourmethod~is developed by a dedicated team of over 20 researchers, including Master's and Ph.D. candidates with expertise in artificial intelligence and related fields. The entire workflow adheres to predefined design principles and a multi-stage quality control pipeline, as delineated in the following subsections.

\subsection{Design Principles}
\ourmethod~follows 3 core design principles that address key limitations of existing benchmarks in task complexity, information searchability, and evaluation dimensions.

\paragraph{\textbf{Multi-dimensional Cross-modal Coverage.}} To more faithfully simulate real-world search scenarios, we augment task complexity along two distinct dimensions. Specifically, we extend search depth via multi-hop variations and categorize cross-modal interaction complexities into 3 hierarchical levels: intra-region alignment, inter-region integration, and inter-image reasoning.

% To faithfully replicate the complexity of real-world search environments, we extend task diversity across two primary axes. First, we vary multi-hop trajectories to ensure depth in the search path. Second, we establish a three-tier hierarchy based on cross-modal interaction complexity: Level 1 (Intra-region Alignment) focuses on fine-grained visual-textual alignment within localized regions; Level 2 (Inter-region Integration) evaluates the synthesis of disparate visual sectors within a single page; and Level 3 (Inter-image Reasoning) challenges the model’s capacity for associative cognition and complex reasoning across multiple distinct images.

\paragraph{\textbf{Process-oriented Granular Evaluation.}} Datasets should incorporate expert-validated sub-goals to enable systematic tracking of intermediate reasoning steps.
This design ensures granular tracking of evidence acquisition phases, thereby permitting a rigorous diagnostic analysis of failure modes and an accurate delineation of model capability boundaries.

% Deviating from traditional end-to-end metrics that focus solely on final outputs, we introduce an expert-verified sub-goal mechanism. By monitoring intermediate reasoning steps, this framework systematically characterizes model performance during evidence acquisition. This approach facilitates not only the quantification of holistic capabilities but also the diagnostic analysis of failure modes, pinpointing specific deficiencies in perception, retrieval, or logical deduction.

\paragraph{\textbf{High Reliability and Reproducibility.}}
For rigorous evaluation, we adopt 3 filtering criteria:
\textbf{(1) Evidence Traceability.} Require all evidence be publicly accessible through search tools with complete manual annotation trajectories.
\textbf{(2) Temporal Stability.} Prioritize temporally invariant, objective knowledge to eliminate dynamic web content fluctuations.
\textbf{(3) Answer Objectivity.} Enforce concise, verifiable answers to enable standardized automated evaluation.

\subsection{Data Construction Pipeline}

% As illustrated in Figure 1, the construction of \ourmethod adheres to a rigorous, closed-loop quality assurance framework. The pipeline is systematically partitioned into 5 distinct s:
As illustrated in Figure~\ref{fig:overview}, \ourmethod~ construction follows a closed-loop quality assurance framework comprising 5 stages:

\paragraph{\textbf{Stage 1: Initialization and Guideline Formulation}} 
The experts defines core evaluation dimensions (domain diversity, task hierarchy, and hop distribution) and constructs Initial Exemplars comprising Visual Inputs, Queries, Sub-goals, Answer, and Metadata. These exemplars, together with Instruction Documents, establish the gold standard for subsequent large-scale annotation.

% \paragraph{\textbf{Stage 1: Initialization and Guideline Formulation}} To guarantee professional rigor and task diversity, the expert team initially defined core evaluation dimensions—including domain distribution, task hierarchy, and hop counts—while synthesizing a suite of high-quality Initial Exemplars. These exemplars comprise comprehensive Visual Inputs, Queries, Sub-goals, Ground-truth Answers, and Metadata, serving as a rigorous benchmark for subsequent annotation. These exemplars, complemented by formal Instruction Documents, establish a ``gold standard'' for large-scale data labeling and provide annotators with a definitive procedural paradigm.

\paragraph{\textbf{Stage 2: Tool-Augmented Exploratory Annotation}}  Annotators are assigned sub-tasks according to domain expertise and conduct exploratory web searches using a suite of specialized tools, including TextSearch, WebVisit, ImageSearch, ImageCrop, and ReverseImageSearch. They document complete interaction trajectories, partition complex tasks into pivotal sub-goals, and annotate the capabilities required to acquire each critical piece of evidence.

% During the annotation phase, the system assigns human annotators to specific sub-tasks based on their domain expertise. Adhering to the established exemplars and guidelines, annotators perform open-ended web browsing utilizing a comprehensive toolkit that integrates TextSearch, WebVisit, ImageCrop, and ReverseImageSearch. The workflow is designed to simulate sophisticated multimodal deep-search scenarios: annotators are required to record complete interaction trajectories while decomposing key sub-goals and identifying the underlying capabilities required for each step. By strategically configuring interaction depth and multi-hop search constraints, this process yields a set of initial QA pairs characterized by significant diversity and search depth.

\paragraph{\textbf{Stage 3: Dual-Verification and Adversarial Filtering}} 
The original dataset undergoes two sequential screening phases. First, in the human verification loop, verifiers replicate the annotated search trajectories and evaluate logical coherence, evidentiary support, and answer accuracy. Samples that fail verification are returned for revision. Second, state-of-the-art (SOTA) multimodal large models filter out trivial examples, ensuring the retention of challenging samples that involve long-tail knowledge or complex reasoning requirements.

% The raw dataset undergoes two rigorous screening phases to ensure the accuracy and complexity of the benchmark. The process begins with a Human Verification Loop, which implements a "Check-Execute-Annotate" (CEA) mechanism. During this phase, independent verifiers replicate the search trajectories to evaluate samples across three dimensions: logical coherence, evidentiary support, and answer accuracy; samples failing to meet these criteria are remanded for remediation. Subsequently, a difficulty-based filtration stage is conducted using state-of-the-art (SOTA) Vision-Language Models (VLMs), such as GPT-4V and Gemini. By filtering out trivial examples that these models solve with ease, this stage preserves highly challenging samples characterized by long-tail distributions or requirements for complex reasoning, thereby enhancing the discriminatory power of the benchmark.

\paragraph{\textbf{Stage 4: Structured Data Formatting}} 
The verified samples are post-processed and converted into a unified JSON format, with standardized input/output fields, sub-goals, and interaction trajectories. This formatting ensures both human readability and machine interpretability, enabling automated evaluation pipelines.
% Following verification, the data proceeds to the format conversion stage, where raw annotation records are cleaned and transformed into a unified structured JSON dataset. This phase standardizes the schema for input and output fields, particularly for sub-goals and fine-grained interaction trajectories. Such formalization ensures both human-readability and machine-interpretability, facilitating seamless integration into automated evaluation pipelines.

\paragraph{\textbf{Stage 5: Expert Quality Control}} Before the formal release, domain experts audit the structured data for safety, privacy compliance, and factual accuracy. Only approved samples are included in the final dataset, ensuring ethical and professional standards.

% Prior to the formal release, the dataset undergoes a terminal quality control phase conducted by domain experts. This panel audits the structured data with a specific focus on data safety, privacy compliance, and the factual accuracy of domain-specific knowledge. Only samples that pass final adjudication are included in the final dataset. This rigorous vetting process ensures that the benchmark adheres to the highest standards of ethical integrity and specialized professional depth.

\begin{figure*}[t]
  \centering
  \begin{minipage}[t]{0.45\textwidth}
    \vspace{0pt}
    \centering
    \includegraphics[width=\linewidth]{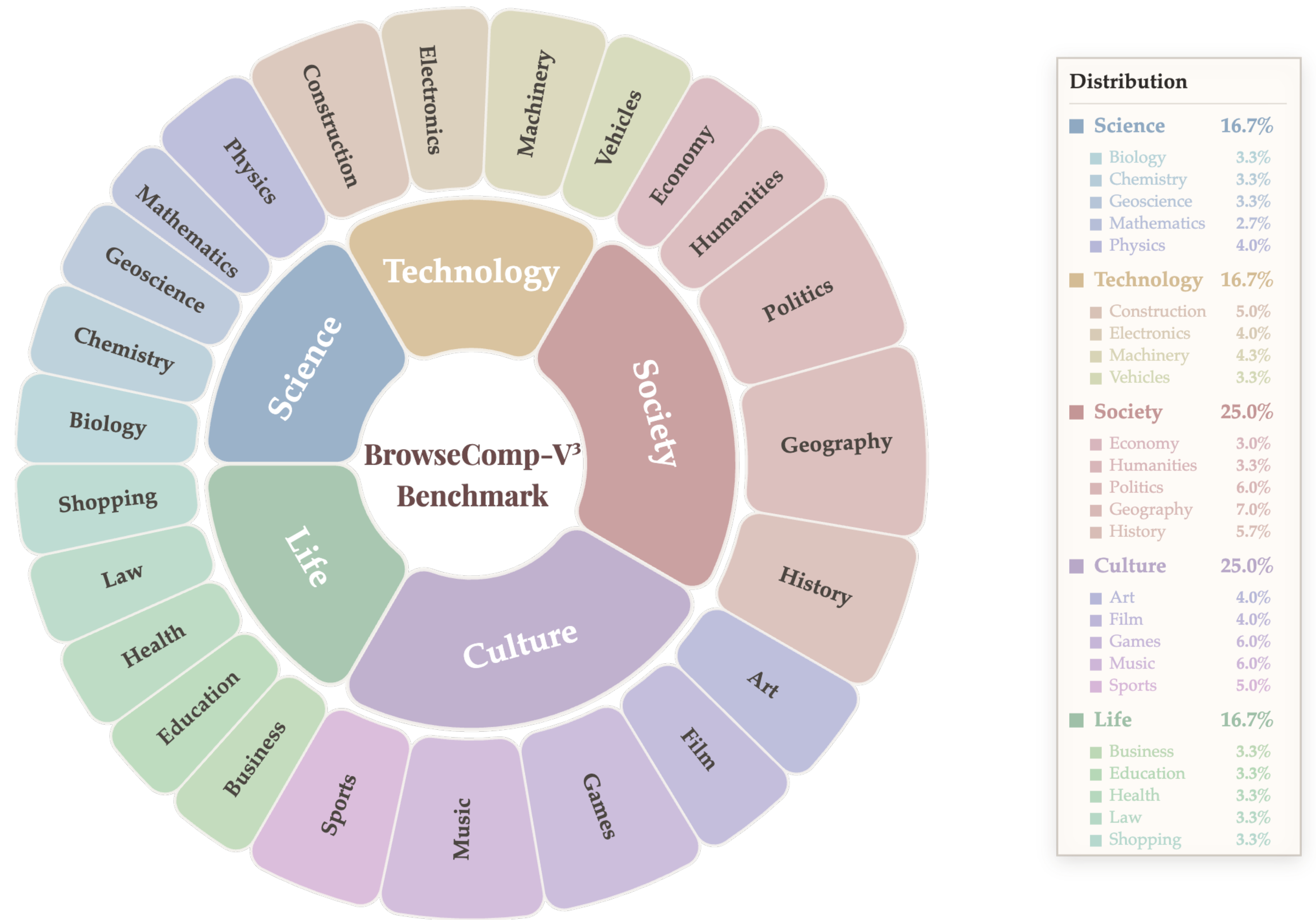}
  \end{minipage}
  \hspace{0.05\textwidth}
  \begin{minipage}[t]{0.4\textwidth}
    \vspace{0pt}
    \centering
    \resizebox{\linewidth}{!}{\input{table/statics}}
  \end{minipage}
  \caption{Statistics of \ourmethod. (Left) Category distribution across primary domains. (Right) Summary of statistics.}
  \label{fig:benchmark}
\end{figure*}

\subsection{Dataset Statistics}
Figure ~\ref{fig:benchmark} (left) illustrates the categorical distribution of~ \ourmethod. The dataset comprises 5 balanced categories: Science, Technology, Society, Culture, and Life. Additional statistical metrics, including basic statistics, task levels and difficulty distributions, are provided in Figure ~\ref{fig:benchmark} (right).
% Figure 2 (a) illustrates the statistical distribution of the \ourmethod dataset. The dataset comprises five evenly distributed categories: Science, Technology, Society, Culture, and Life. To encompass authentic multimodal deepsearch scenarios, we established a hierarchical structure based on information interaction complexity, consisting of three progressive levels: Level 1 (Intra-regional alignment), Level 2 (Cross-regional integration), and Level 3 (Cross-image reasoning). Furthermore, to facilitate a multi-dimensional evaluation of model performance, we defined four difficulty tiers—ranging from Easy to Hard—based on the number of search hops. Detailed statistical metrics are provided in Figure 2 (b).

\section{Experiments}

\subsection{Experimental Setup}

\paragraph{\textbf{Evaluated Models}} We systematically evaluate \ourmethod under 4 representative settings, as detailed below:

\begin{itemize}[leftmargin=0.5cm] 
    \item \textbf{Human.} To assess human performance, we recruit participants with PhD-level expertise who independently solve each problem utilizing a standard web browser. Participants can freely browse publicly accessible web resources to gather evidence and produce verifiable answers.
    \item \textbf{Tool-Free MLLMs.} We benchmark multiple SOTA MLLMs in a tool-free setting, where models must generate answers directly without access to external tools or search capabilities. Specifically, we evaluate the following models: GPT-5.2~\citep{openai2025gpt52}, o4-mini~\citep{openai2025o3o4mini}, GPT-4o~\citep{Hurst2024GPT4oSC}, Gemini-3-Flash-Preview~\citep{google2025gemini3flash}, Claude-Sonnet-4.5~\citep{anthropic2025claudesonnet45}, Doubao-Seed-1.8~\citep{bytedance2025seed18}, MiMo-V2-Flash~\citep{coreteam2026mimov2flashtechnicalreport}, Qwen3-VL-235B-A22B-Instruct~\citep{bai2025qwen3vltechnicalreport}, and Qwen3-VL-8B-Instruct~\citep{bai2025qwen3vltechnicalreport}.
    \item \textbf{Tool-Augmented MLLMs.} Additionally, we evaluate tool-augmented model services accessed through their official web platforms, with the maximum reasoning mode enabled to elicit their full capabilities. Concretely, we evaluate the following models: GPT-5.2-Thinking~\citep{gptweb}, Gemini-3-Pro-Preview~\citep{geminiweb}, and Claude-Sonnet-4.5-Thinking~\citep{claudeweb}.
    \item \textbf{OmniSeeker.} Lastly, we evaluate models using OmniSeeker, our custom-built multimodal browsing agent—a unified and transparent framework equipped with standardized tools, including TextSearch, WebVisit, ImageSearch, ImageCrop, and ReverseImageSearch.
\end{itemize}

\paragraph{\textbf{Implementation Details}} 
We employ a unified and rigorous evaluation protocol across all four settings. For the \textbf{human baseline}, participants have up to 30 minutes per question; if they cannot reach a reliable conclusion within this time limit, they may terminate the task and document their key exploration steps. For \textbf{Tool-Augmented MLLMs}, we enable the most advanced reasoning mode available to ensure unconstrained model performance. For \textbf{Tool-Free MLLMs}, models receive only the question text and original images without any tool access, and must directly generate the key information and final answer. Under the \textbf{OmniSeeker} setting, we limit interactions to a maximum of 20 rounds per question. The retrieval module uses Serper\footnote{\url{https://serper.dev}} and returns the top 5 results; image retrieval outputs are embedded into the dialogue context as base64-encoded data; the webpage access module uses Jina\footnote{\url{https://jina.ai}} to retrieve and parse webpage content; and image cropping is performed programmatically, with cropped images returned to the model.

\paragraph{\textbf{Evaluation Metrics}} We employ both result-level and process-level metrics. At the result level, we use \textbf{Success Rate} to measure whether tasks are completed successfully. At the process level, we introduce \textbf{Process Score} to quantify how much progress a model makes toward problem resolution during multi-step search and reasoning—specifically, the proportion of critical sub-goals successfully completed. This metric is formally defined as:

\begin{equation}
    \mathrm{ProcessScore}(q) = \frac{| \hat{\mathcal{G}}_q|}{|\mathcal{G}_q|},
\end{equation}

where $\mathcal{G}_q$ denotes the set of ground-truth sub-goals required to solve problem $q$, and $\hat{\mathcal{G}}_q$ denotes the set of sub-goals achieved by the model or human.

\input{table/main_result}
\subsection{Main Results}

Based on the experimental results in Table \ref{tab: main_result}, we summarize our key findings as follows:

\textbf{(1) Performance Gap and Benchmark Difficulty.} Humans significantly outperform all models on \ourmethod tasks, achieving an average Success Rate of $68.03\%$ and a Process Score of $82.93\%$. In contrast, no model achieves more than $40\%$ SR. This gap both highlights the limitations of current MLLMs in multimodal deep search tasks and validates the benchmark's ability to capture real-world search complexity.

\textbf{(2) Critical Role of Tool Augmentation.} Without tool access, most models achieve only approximately $10\%$ SR. Tool augmentation yields substantial performance improvements, indicating that parameterized knowledge alone cannot adequately capture dynamic, cross-modal evidence chains on the open web. This highlights the importance of external retrieval and interactive capabilities for deep multimodal reasoning.

\textbf{(3) Effectiveness and Generalizability of OmniSeeker.} Empirical evidence confirms that OmniSeeker provides a unified and efficient tool-calling framework. When equipped with OmniSeeker, all models consistently achieve substantial improvements, reaching performance comparable to specialized proprietary systems.

\textbf{(4) Value of Process-Level Evaluation.} We observe a notable gap between PS and SR, with PS typically exceeding SR. This indicates that while models can complete individual sub-goals, they often fail to maintain logical consistency across long-sequence tasks. Therefore, fine-grained process-level evaluation is essential for identifying where and why models fail, thereby revealing their capability boundaries.

\textbf{(5) Competitive Performance of Open-Source Models.} While proprietary models (e.g., GPT-5.2~\citep{openai2025gpt52}) remain the top performers, high-performance open-source models are rapidly closing the gap. Notably, Doubao-Seed-1.8~\citep{bytedance2025seed18} achieves $33.67\%$ SR when equipped with OmniSeeker. This demonstrates that high-quality open-source models possess strong capacity for complex reasoning and provide a promising path toward developing cost-effective, high-performance web browsing agents.

\input{table/level_new}

\section{Further Analysis}

\subsection{Fine-grained Analysis}

\paragraph{\textbf{Task Level}} As shown in Table~\ref{tab:level}, model performance declines substantially as task complexity increases from Level 1 to Levels 2 and 3. This reveals that while models can effectively perform unitary visual search, they face significant challenges in inter-region integration and inter-image relational reasoning.

\begin{figure*}[t]
\centering
\includegraphics[width=0.45\linewidth]{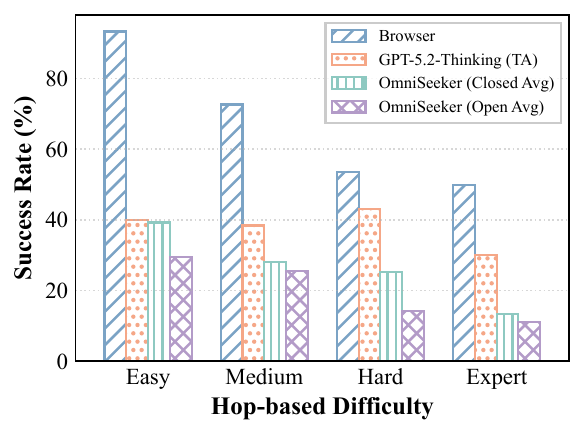}\hfill
\includegraphics[width=0.45\linewidth]{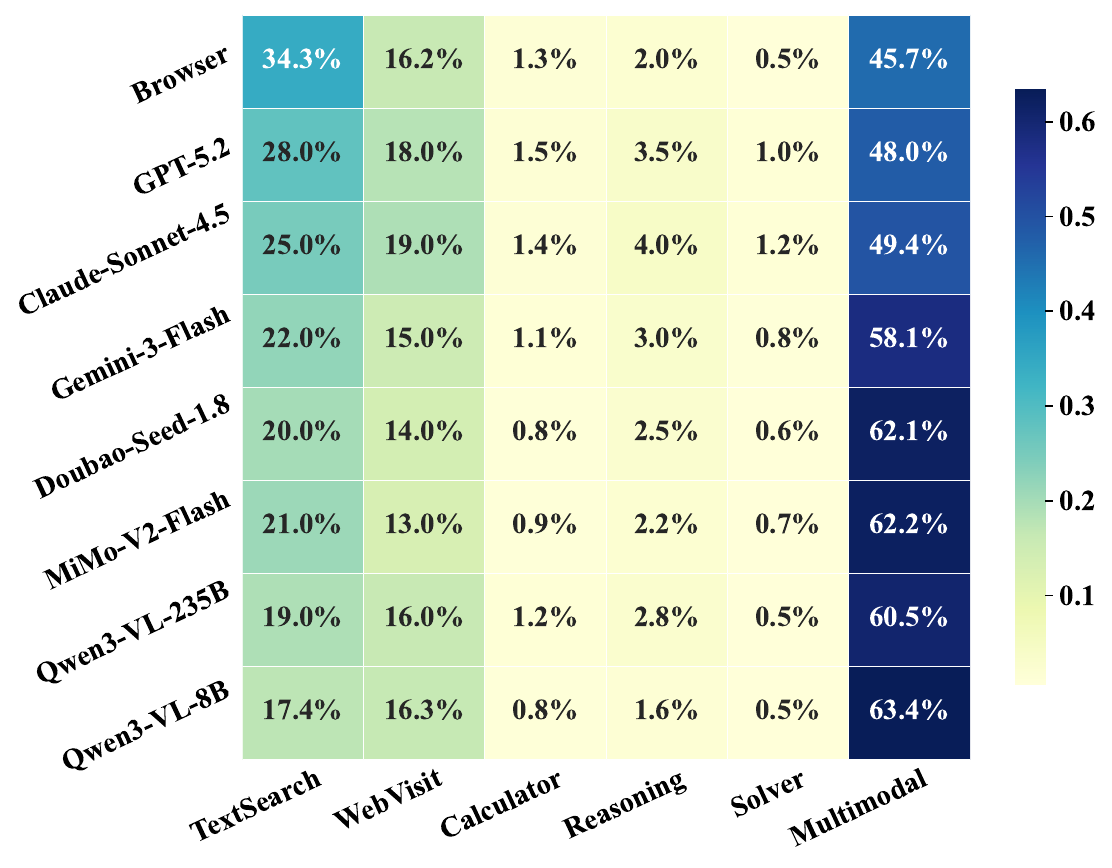}
\caption{Difficulty and Ability Analysis}
\label{fig:diff_ability}
\end{figure*}

\paragraph{\textbf{Search Depth}} As illustrated in Figure~\ref{fig:diff_ability} (Left), SR for both humans and models decline with increasing search depth, yet exhibit distinct patterns. Human performance drops sharply with longer search paths, whereas model performance declines more gradually. This discrepancy implies that models leverage internalized parametric knowledge as a compensatory mechanism to mitigate the impact of search complexity.

\paragraph{\textbf{Ability Boundaries}} Figure~\ref{fig:diff_ability} (Right) reveals distinct bottlenecks for humans and models. Human performance limitations are primarily in TextSearch, due to constraints in attention span and cognitive load when processing voluminous text. In contrast, multimodal integration remains the primary bottleneck for all models.

\begin{figure*}[t]
\centering
\includegraphics[width=0.45\linewidth]{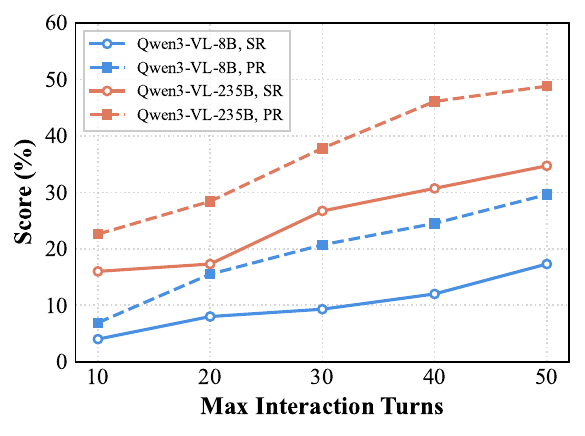}\hfill
\includegraphics[width=0.45\linewidth]{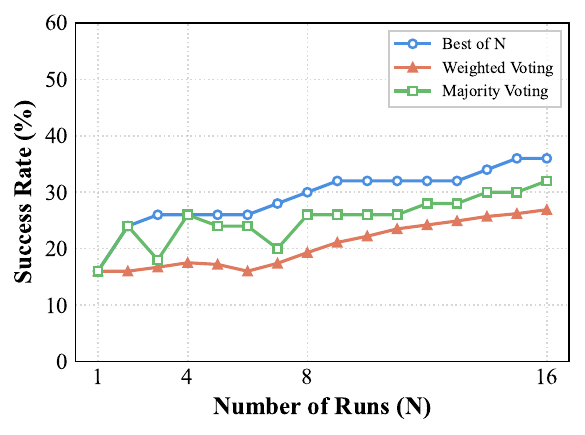}
\caption{Test Time Scaling}
\label{fig:test}
\end{figure*}

\begin{figure*}
    \centering
    \includegraphics[width=\textwidth]{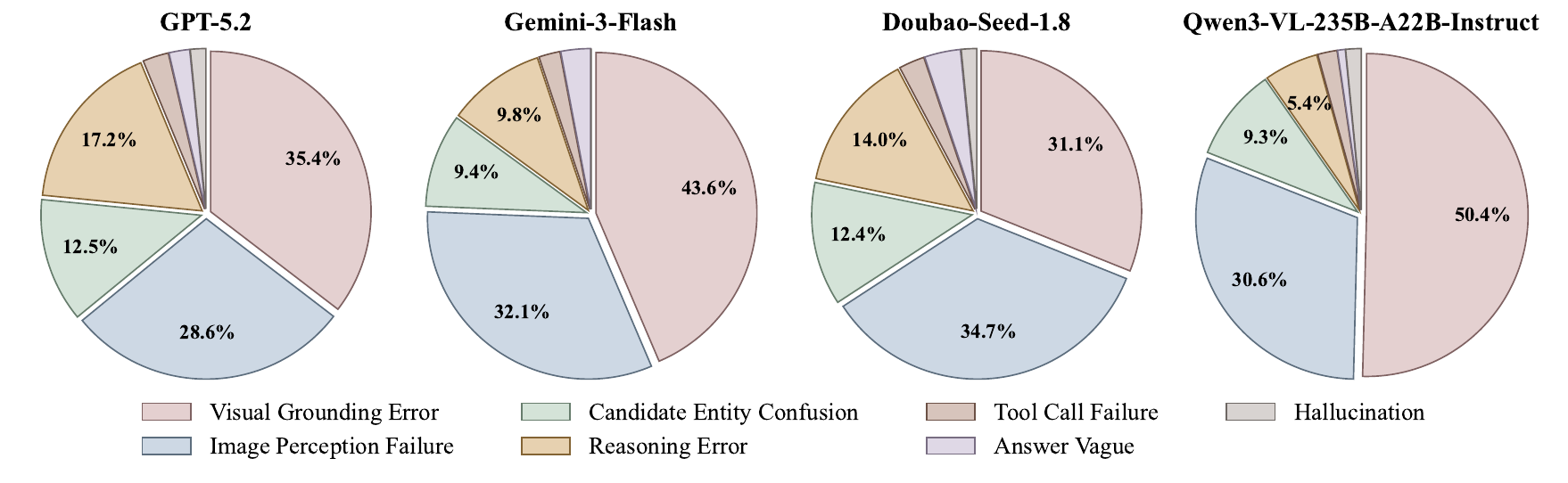}
    \setlength{\abovecaptionskip}{1pt}
    \caption{Failure Mode Analysis}
    \label{fig:fail}
\end{figure*}
\subsection{Test Time Scaling}
We evaluate how test-time compute affects performance on \ourmethod. Our key findings are as follows:

\begin{itemize}[leftmargin=0.5cm] 
    \item \textbf{Scaling Interaction Steps.} As shown in Figure~\ref{fig:test} (Left), increasing the maximum number of interaction turns substantially improves performance. Notably, Qwen3-VL-235B exhibits a stronger scaling advantage than its 8B counterpart. This suggests that larger models have stronger long-horizon reasoning capabilities, allowing them to better utilize additional interaction steps for iterative refinement.
    \item \textbf{Scaling Sampling Consistency.} Figure~\ref{fig:test} (Right) shows the performance of Qwen3-VL-235B as we increase the number of independent samples ($N$). Among the three strategies, Best-of-$N$ scales most effectively, continuously improving performance with increasing $N$. 
    % In contrast, while Weighted Voting and Majority Voting yield initial improvements, their gains diminish and eventually plateau.
\end{itemize}

\subsection{Failure Mode Analysis}

We analyze the error distributions of four representative models, as shown in Figure~\ref{fig:fail}. Our key findings are summarized below:

\begin{itemize}[leftmargin=0.5cm] 
    \item \textbf{Multimodal Grounding and Perception.} Across all models, Visual Grounding and Perception Failure dominate the error distribution. This indicates that current MLLMs struggle to accurately extract and perceive visual information in complex, noisy web environments.
    \item \textbf{Multimodal Progress, Planning Constraint.} Closed-source models substantially reduce perception and grounding errors compared to open-source models. However, with improved multimodal capabilities, long-horizon planning becomes the main bottleneck limiting further improvements in SOTA models.
\end{itemize}

\section{Conclusion}
In this work, we introduce \ourmethod, a comprehensive benchmark for the evaluation of multimodal deep browsing and search capabilities. 
The benchmark consists of 300 rigorously curated and annotated questions designed to systematically remedy three core limitations of existing evaluation paradigms: task complexity, information searchability, and evaluation dimensions. 
Empirical results reveal that SOTA MLLMs achieve under 40\% SR, underscoring a substantial gap relative to human performance. 
These findings confirm the effectiveness and discriminative power of \ourmethod in simulating open-world multimodal deep search scenarios. Further analysis reveals critical deficiencies in current models' capacity to integrate and comprehend multimodal information, whereas the process-level evaluation and Test-Time Scaling analysis offer potential pathways for enhancing model capabilities via methodologies such as reinforcement learning. 
Additionally, our agent framework, OmniSeeker, achieves performance comparable to leading closed-source models, offering an open alternative for developing multimodal browsing agents.
In conclusion, \ourmethod provides a comprehensive platform for evaluating and advancing multimodal browsing agents.
Its process-level evaluation and fine-grained capability analysis will catalyze future breakthroughs in multimodal deep search.

%%%%%%%%%%%%%%%%%%%%%%%%%%%%%%%%%%%%%%%%%%%%%%%%%%%%%%%%%%

%%%%%%%%%%%%%%%%%%%%%%%%%%%%%%%%%%%%%%%%%%%%%%%%%%%%%%%%%%%%

% \clearpage
% \bibliographystyle{unsrtnat}

%%%%%%%%%%%%%%%%%%%%%%%%%%%%%%%%%%%%%%%%%%%%%%%%%%%%%%%%%%%%

%%%%%%%%%%%%%%%%%%%%%%%%%%%%%%%%%%%%%%%%%%%%%%%%%%%%%%%%%%%%

\newpage

\end{document}

%% file: table/comparison.tex
\begin{table*}[t]
    \centering
    \caption{Comparison of our benchmark against representative deep search benchmarks along eight dimensions. $^\S$ indicates that the benchmark only partially satisfies the corresponding criterion.}
    % \small
    \renewcommand{\arraystretch}{1.3}
    \resizebox{\textwidth}{!}{%
    \begin{tabular}{lcccccccc}
        \toprule
        \textbf{Benchmarks} &
        \makecell{\textbf{Multimodal }\\\textbf{Context inputs}} &
        \makecell{\textbf{Multi-round}\\\textbf{Interaction(>2)}} &
        \makecell{\textbf{Thinking with}\\\textbf{Images}} &
        \makecell{\textbf{Multi-image}\\\textbf{Reasoning}} & \makecell{\textbf{Public-search}\\\textbf{Answerable}} &
        \makecell{\textbf{Hop-based}\\\textbf{Difficulty Analysis}} & \makecell{\textbf{Human-validated}\\\textbf{Trajectories
}} & \makecell{\textbf{Fine-grained}\\\textbf{Progress Metrics}}  
        \\
        \midrule
        InfoSeek~\cite{chen2023infoseek}  & \cmark          & \xmark & \xmark & \xmark & \cmark & \xmark & \xmark & \xmark\\
        Enc-VQA~\cite{mensink2023Enc-VQA} & \cmark          & \xmark & \xmark & \xmark & \cmark & \xmark & \xmark & \xmark\\
        MMSearch~\cite{jiang2024mmsearch} & \cmark          & \xmark & \xmark & \xmark & \cmark & \xmark & \xmark & \xmark\\
        DynVQA~\cite{li2024DynVQA} & \cmark & \xmark$^\S$ & \xmark & \xmark & \cmark & \cmark & \xmark & \xmark\\
        SimpleVQA~\cite{cheng2025simplevqa}  & \cmark  & \xmark & \xmark & \xmark & \cmark & \xmark & \xmark & \xmark\\
        LiveVQA~\cite{fu2025livevqa}  & \cmark & \xmark$^\S$ & \xmark & \xmark & \xmark$^\S$ & \cmark & \xmark & \xmark\\
        BrowseComp~\cite{wei2025browsecomp}    & \xmark & \cmark & \xmark & \xmark & \cmark & \xmark & \xmark & \xmark\\
        FactualVQA~\cite{wu2025mmsearchr1} & \cmark & \xmark & \xmark & \xmark & \cmark & \xmark & \xmark & \xmark \\
        
        BrowseComp-VL~\cite{geng2025webwatcher}  & \cmark & \cmark & \xmark & \xmark & \cmark & \xmark & \xmark & \xmark \\
        MM-BrowseComp~\cite{li2025mmbrowsecomp}  & \cmark & \cmark & \cmark & \xmark & \xmark$^\S$ & \xmark & \xmark & \xmark \\
        MMSearch-Plus~\cite{tao2025mmsearchplus} & \cmark & \cmark & \cmark & \cmark & \xmark$^\S$ & \xmark & \xmark & \xmark \\
        \midrule
        \textbf{\ourmethod~(Ours)}           & \cmark & \cmark & \cmark & \cmark & \cmark & \cmark & \cmark & \cmark \\
        \bottomrule
    \end{tabular}}
\end{table*}

%% file: table/statics.tex
\resizebox{0.9\textwidth}{!}{
\begin{tabular}{llr}
\toprule
\textbf{Type} & \textbf{Statistic} & \textbf{Number} \\
\midrule
\multirow{6}{*}{Basic Statistics} 
    & Total questions & 300 \\
    & Total images & 383 \\
    & Maximum question length & 134 \\
    & Maximum answer length & 23 \\
    & Average question length & 58.58 \\
    & Average answer length & 2.47 \\
\midrule
\multirow{2}{*}{Category Statistics}
    & Primary & 5 \\
    & Secondary & 24 \\
\midrule
\multirow{3}{*}{Task Level}
    & Level 1 & 89 \\
    & Level 2 & 140 \\
    & Level 3 & 71 \\
\midrule
\multirow{4}{*}{Difficulty Distribution}
    & Easy & 45 \\
    & Medium & 139 \\
    & Hard & 86 \\
    & Expert & 30 \\
\bottomrule
\end{tabular}
}

%% file: table/main_result.tex
\begin{table*}[t]
\centering
\caption{
\textbf{Performance on \ourmethod.}
Results are reported in terms of \textbf{Success Rate} and \textbf{Process Score} under the Pass@1 setting.
\textbf{Avg.} denotes the average performance, while \textbf{Sci.}, \textbf{Tech.}, \textbf{Soc.}, \textbf{Cul.}, and \textbf{Lif.} correspond to the \emph{Science}, \emph{Technology}, \emph{Society}, \emph{Culture}, and \emph{Life} categories, respectively.
Bold numbers indicate the best-performing model within each group.
}
\renewcommand{\arraystretch}{1.2}
\resizebox{\textwidth}{!}{
% The column definition adds extra horizontal space between the OA and SA groups for readability
% \setlength{\tabcolsep}{3pt}
\begin{tabular}{l|>{\columncolor{lightgray}}{c}|ccccc|>{\columncolor{lightgray}}{c}|ccccc}
\toprule
\multirow{2}{*}{\textbf{Model}} 
& \multicolumn{6}{c}{\textbf{Success Rate (SR, \%)}} 
& \multicolumn{6}{c}{\textbf{Process Score (PS, \%)}} \\
\cmidrule(lr){2-7} \cmidrule(lr){8-13}
& \textbf{Avg.} & \textbf{Sci.} & \textbf{Tech.} & \textbf{Soc.} & \textbf{Cul.} & \textbf{Lif.}
& \textbf{Avg.} & \textbf{Sci.} & \textbf{Tech.} & \textbf{Soc.} & \textbf{Cul.} & \textbf{Lif.} \\
\midrule

\multicolumn{13}{c}{\textbf{Human}} \\
\midrule
Browser & \textbf{68.03} & 72.00 & 70.00 & 73.33 & 68.00 & 54.00 & \textbf{82.93} & 87.54 & 85.25 & 84.19 & 82.73 & 74.32 \\
\midrule

\multicolumn{13}{c}{\textbf{Tool-Augmented MLMs}} \\
\midrule
GPT-5.2-Thinking & \textbf{39.13} & 26.00 & 48.00 & 38.67 & 37.33 & 46.00 & \textbf{66.05} & 61.11 & 79.74 & 54.87 & 71.41 & 64.73 \\
Gemini-3-Pro-Preview & 22.90 & 18.00 & 16.00 & 21.33 & 24.00 & 34.00 & 62.43 & 62.02 & 73.78 & 48.70 & 66.33 & 62.50 \\
Claude-Sonnet-4.5-Thinking & 18.33 & 22.00 & 16.00 & 17.33 & 21.33 & 14.00 & 47.73 & 58.61 & 58.41 & 34.79 & 54.33 & 35.68 \\
\midrule

\multicolumn{13}{c}{\textbf{Tool-Free MLMs}} \\
\midrule
GPT-5.2 & 6.00 & 0.00 & 14.00 & 4.00 & 5.33 & 8.00 & 25.02 & 17.50 & 44.50 & 19.22 & 25.25 & 21.43 \\
o4-mini & 7.33 & 0.00 & 16.00 & 2.67 & 6.67 & 14.00 & 29.08 & 29.48 & 46.12 & 17.91 & 29.04 & 28.44 \\
GPT-4o & 2.67 & 0.00 & 10.00 & 2.67 & 1.33 & 0.00 & 11.26 & 6.78 & 26.52 & 9.45 & 7.60 & 8.66 \\
Gemini-3-Flash-Preview & \textbf{12.00} & 8.00 & 18.00 & 12.00 & 10.67 & 12.00 & \textbf{40.76} & 38.98 & 61.94 & 31.89 & 39.72 & 36.60 \\
Claude-Sonnet-4.5 & 4.00 & 4.00 & 6.00 & 2.67 & 5.33 & 2.00 & 25.74 & 33.06 & 47.56 & 15.97 & 24.53 & 13.06 \\
Doubao-Seed-1.8 & 9.00 & 8.00 & 16.00 & 1.33 & 13.33 & 8.00 & 34.74 & 36.26 & 51.00 & 22.28 & 39.87 & 27.96 \\
MiMo-V2-Flash & 3.00 & 2.00 & 4.00 & 2.67 & 4.00 & 2.00 & 8.12 & 4.52 & 17.28 & 5.43 & 9.40 & 4.70 \\
Qwen3-VL-235B-A22B-Instruct & 3.33 & 4.00 & 4.00 & 4.00 & 2.67 & 2.00 & 20.52 & 26.34 & 35.38 & 11.69 & 19.64 & 14.39 \\
Qwen3-VL-8B-Instruct & 1.00 & 0.00 & 2.00 & 1.33 & 0.00 & 2.00 & 6.64 & 3.28 & 15.36 & 6.39 & 4.09 & 5.48 \\
\midrule

\multicolumn{13}{c}{\textbf{OmniSeeker (Ours)}} \\
\midrule
GPT-5.2 & \textbf{36.00} & 50.00 & 28.00 & 33.33 & 33.33 & 38.00 & 57.70 & 67.23 & 55.40 & 49.34 & 60.49 & 58.81 \\
o4-mini & 26.00 & 22.00 & 24.00 & 25.33 & 34.67 & 20.00 & 44.66 & 43.70 & 52.11 & 40.73 & 48.72 & 37.94 \\
GPT-4o & 11.41 & 14.00 & 14.00 & 10.67 & 14.67 & 2.00 & 24.15 & 22.32 & 43.36 & 17.35 & 25.93 & 13.04 \\
Gemini-3-Flash-Preview & 23.67 & 32.00 & 24.00 & 18.67 & 25.33 & 20.00 & 47.37 & 50.84 & 68.35 & 40.15 & 43.68 & 39.27 \\
Claude-Sonnet-4.5 & 22.67 & 32.00 & 20.00 & 21.33 & 24.00 & 16.00 & 54.17 & 60.94 & 64.25 & 45.29 & 56.73 & 46.78 \\
Doubao-Seed-1.8 & 33.67 & 42.00 & 28.00 & 37.33 & 38.67 & 18.00 & \textbf{58.44} & 52.94 & 69.70 & 57.46 & 66.27 & 42.42 \\
MiMo-V2-Flash & 16.67 & 18.00 & 12.00 & 10.67 & 29.33 & 10.00 & 31.33 & 34.84 & 35.04 & 24.45 & 42.43 & 17.76 \\
Qwen3-VL-235B-A22B-Instruct & 14.33 & 16.00 & 14.00 & 14.67 & 17.33 & 8.00 & 26.68 & 28.56 & 35.93 & 22.00 & 28.36 & 20.05 \\
Qwen3-VL-8B-Instruct & 5.33 & 2.00 & 2.00 & 9.33 & 8.00 & 2.00 & 13.40 & 8.02 & 19.30 & 15.57 & 15.55 & 6.38 \\

\bottomrule
\end{tabular}
}
\label{tab: main_result}
\end{table*}

%% file: table/level_new.tex
\begin{wraptable}{l}{0.50\columnwidth}
  \vspace{-4pt}
  \centering
  \caption{PS across Different Models and Levels}
  \label{tab:level}
  \renewcommand{\arraystretch}{1.1}
  \begin{tabular}{l|ccc}
    \hline
    \textbf{Model} & \textbf{L1} & \textbf{L2} & \textbf{L3} \\
    \hline
    GPT-5.2 & 0.6176 & 0.5528 & 0.5792 \\
    Claude-Sonnet-4.5 & 0.5708 & 0.5353 & 0.5186 \\
    Doubao-Seed-1.8 & 0.6185 & 0.5652 & 0.5838 \\
    MiMo-V2-Flash & 0.3776 & 0.2638 & 0.3420 \\
    Qwen3-VL-235B & 0.3262 & 0.2308 & 0.2715 \\
    \hline
  \end{tabular}
  \vspace{-8pt}
\end{wraptable}